\documentclass[12pt]{article}

\usepackage[margin=1.25in]{geometry}
\usepackage{times}
\usepackage{latexsym}
\usepackage[T1]{fontenc}
\usepackage[utf8]{inputenc}
\usepackage{microtype}
\usepackage{inconsolata}
\usepackage{graphicx}
\usepackage{natbib}
\usepackage{url}
\usepackage{authblk}

\usepackage{algorithm}
\usepackage{algorithmic}
\usepackage{algorithm}
\usepackage{algorithmic}
\usepackage{amsfonts}
\usepackage{amsmath}
\usepackage{bm}

\usepackage{booktabs}
\usepackage{subcaption}
\usepackage{float}
\usepackage{multirow}

\newcommand{\customsmall}{\fontsize{8}{10}\selectfont}

\title{\Large Leveraging LLMs for Early Alzheimer's Prediction}

\author{Tananun Songdechakraiwut \\
  {\small Department of Computer Science\\Duke University}}
\date{}

\begin{document}

\maketitle

\begin{abstract}
We present a connectome-informed LLM framework that encodes dynamic fMRI connectivity as temporal sequences, applies robust normalization, and maps these data into a representation suitable for a frozen pre-trained LLM for clinical prediction. Applied to early Alzheimer's detection, our method achieves sensitive prediction with error rates well below clinically recognized margins, with implications for timely Alzheimer's intervention.
\end{abstract}

\section{Introduction}

Early and accurate detection of neurological and psychiatric disorders remains a major challenge in clinical neuroscience \citep{laske2015innovative}. Brain connectomics offers promise for identifying biomarkers sensitive to early disease processes \citep{woo2017building,yi2025topological}. However, translating complex, high-dimensional, noisy, and time-varying connectivity patterns into findings with clinical utility is difficult, particularly given the prevalence of small and imbalanced labeled datasets.

Recent advances in large language models (LLMs) have opened new opportunities for clinical neuroimaging \citep{brown2020language}. LLMs can capture complex dependencies in sequential data and generalize across tasks by drawing on extensive pre-trained knowledge~\citep{jin2024timellm,varadarajan2025augmenting}. However, adapting these models for the unique challenges of clinical brain connectivity data requires careful consideration of data representation, normalization, and model robustness.

To address these challenges, we present a connectome-informed LLM framework for clinical prediction tasks. Each patient's dynamic brain connectivity is represented as a temporal sequence derived from fMRI data, then normalized to address potential artifacts and outliers common in clinical time series. These normalized sequences are embedded and transformed into a format compatible with a frozen LLM, enabling data processing without modifying the model's core parameters. During training, only lightweight, task-specific modules are updated, allowing effective learning from limited clinical data while leveraging the pre-trained LLM's capabilities.

We apply this framework to early detection of Alzheimer's disease, evaluating its performance on the OASIS-3 dataset \citep{lamontagne2019oasis} with limited cases of early cognitive impairment, reflecting the rarity typical in real-world clinical populations. Our results show sensitive prediction of clinical severity and early impairment, with error rates well below clinically recognized margins, highlighting the framework's potential utility for clinical neuroimaging.

\section{Connectome-LLM Framework}

\begin{figure}
  \centering
  \includegraphics[width=.65\columnwidth]{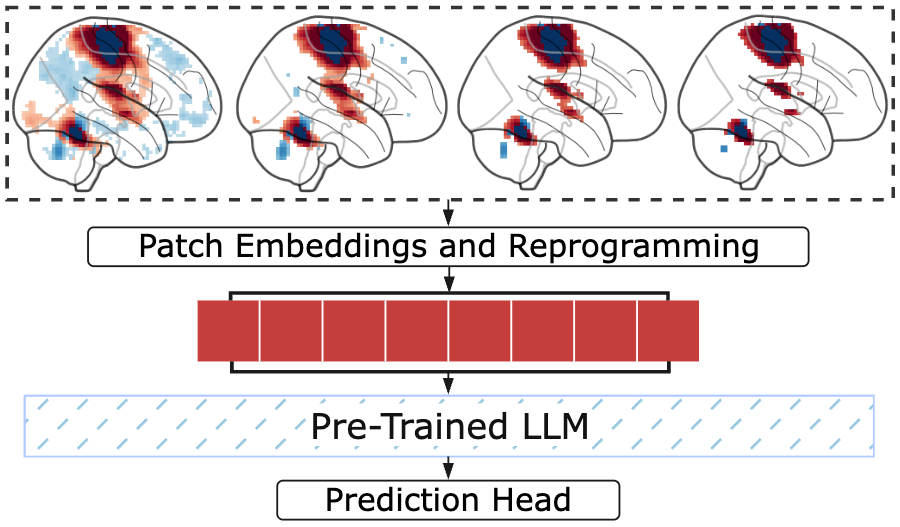}
  \caption{Overview of connectome-LLM framework.}
  \label{fig:schematic}
\end{figure}

A schematic of our framework is shown in Figure~\ref{fig:schematic}. Each subject is encoded as a temporal sequence of dynamic functional connectomes extracted over time. The model takes these sequences as input and predicts clinically relevant targets, such as diagnostic status or symptom severity. During training, only the task-specific modules are updated, with the LLM backbone kept frozen.

\subsection{Dynamic functional connectomes}

Our framework begins with fMRI data, which records blood oxygenation level-dependent (BOLD) signal fluctuations across the brain over time \citep{caballero2017methods}. To summarize this complex spatiotemporal data, we first parcellate the brain into regions of interest (ROIs) using an anatomical or functional brain atlas \citep{bullmore2009complex}, and extract the mean BOLD time series for each region.

Dynamic functional connectivity is then computed by estimating, for each sliding time window, a connectivity matrix that reflects the statistical relationships (such as Pearson correlation) between ROI pairs within that window \citep{songdechakraiwut2020dynamic}. Shifting the window across the scan produces a time-ordered sequence of connectivity matrices for each subject, encoding the evolving patterns of functional coupling between regions.

To capture the temporal dynamics of these networks, our framework computes the difference between each pair of consecutive connectivity matrices using a set of distance functions, each designed to capture distinct aspects of network topology or geometry \citep{rubinov2010complex}. In practice, these distance functions can include standard global or local network metrics. While data-driven approaches such as graph neural networks can be used to learn representations from connectivity data \citep{hamilton2020graph}, in data-limited medical settings we prioritize well-established, interpretable metrics to mitigate overfitting and ensure model transparency.

This process transforms the sequence of connectivity matrices into a univariate or multivariate time series of network change scores. Each entry in this series reflects the magnitude and nature of topological change in functional connectivity from one window to the next, providing a compact and interpretable description of brain network dynamics.

\subsection{Temporal patch embedding}

The multivariate time series derived from dynamic functional connectomes, denoted as $\mathbf{X} \in \mathbb{R}^{d \times T}$, where $d$ is the number of features and $T$ is the number of time windows, serves as the input for downstream representation learning. Each feature sequence, $\mathbf{X}^{(i)} \in \mathbb{R}^{1 \times T}$ for $i=1,\ldots,d$, is normalized using \emph{reversible robust instance normalization} (RevRIN). For each sequence, the median is subtracted and the result is divided by the interquartile range. RevRIN reduces the influence of outliers and spikes in brain activity, which are common in fMRI due to artifacts or transient neural events, and improves stability when the data distribution deviates from Gaussianity or contains abrupt fluctuations. The original scale of each feature can be restored after model inference by applying the inverse transformation. 

To further structure the input for learning temporal patterns, each normalized sequence is divided into overlapping temporal patches \citep{nie2022time} using a sliding window of length $L$ and stride $S$. This process produces, for each feature $i$, a set of $m$ patches $\tilde{\mathbf{X}}^{(i)} \in \mathbb{R}^{m \times L}$, where $m = \left\lfloor \frac{T-L}{S} \right\rfloor + 1$. Each patch captures the local temporal context of the connectivity dynamics for that feature.

Each temporal patch is then projected into a low-dimensional embedding space through a learnable transformation, resulting in an embedded representation $\hat{\mathbf{X}}^{(i)} \in \mathbb{R}^{m \times d_e}$, where $d_e$ is the embedding dimension. The collection of these embedded patches across all features provides a compact and structured summary of temporal dynamics, forming the input to the subsequent LLM.

\subsection{Patch-to-token reprogramming}

After temporal patch embedding, the embedded patches $\hat{\mathbf{X}}^{(i)} \in \mathbb{R}^{m \times d_e}$ for each feature $i$ are mapped into the input space of the LLM through a patch-to-token reprogramming module \citep{jin2024timellm}.

In this step, a learnable transformation projects each patch embedding to the hidden dimension of the frozen LLM, denoted $d_h$. To ensure semantic compatibility with the LLM, we employ a multi-head cross-attention mechanism that aligns each patch embedding with a fixed set of semantic anchors. These anchors are defined as a subset of token embeddings selected from the LLM's vocabulary and serve as text prototypes, providing a bridge between the patch representations and the language model's input space. The output of this reprogramming step is a sequence $\mathbf{Z}^{(i)} \in \mathbb{R}^{m \times d_h}$, where each patch is now represented as a token in the LLM's input format, preserving temporal context and enabling integration with the model's pretrained architecture.

By leveraging these semantic anchors, the patch-to-token mapping enables the model to reason over dynamic connectome features using the expressive capacity and pretrained knowledge of the LLM. The reprogrammed token sequences for all features are concatenated to form the complete LLM input for downstream regression or classification tasks.

\subsection{LLM-based prediction}

The reprogrammed token sequences generated from all features are concatenated to form the final input to the LLM. The frozen LLM processes this input sequence, generating contextualized representations that capture both local and global dependencies across the dynamic functional connectome.

For each input sequence, the LLM produces a sequence of hidden states $\mathbf{H} \in \mathbb{R}^{M \times d_h}$, where $M = d \times m$ is the total number of reprogrammed tokens (with $d$ the number of features and $m$ the number of patches per feature). A lightweight output head, such as a linear projection applied to a flattened version of $\mathbf{H}$, produces the final prediction $\hat{y}$. This design allows the model to flexibly output regression targets or classification probabilities, depending on the downstream task.

\subsection{Optimizing for clinical imbalance}

Our framework is trained in an end-to-end fashion for regression or classification tasks on dynamic functional connectome data. The objective function is selected according to the downstream application, with mean squared error used for regression or cross-entropy loss for classification.

To address class imbalance and the rarity of positive diagnostic cases in clinical datasets, we employ an outlier-weighted mean squared error (OW-MSE) loss during training. This loss is defined as $\mathcal{L}_{\mathrm{OW\text{-}MSE}} = \frac{1}{N} \sum_{i=1}^N w_i (\hat{y}_i - y_i)^2$, where $\hat{y}_i$ and $y_i$ denote the predicted and true target for sample $i$, and $w_i$ is a sample weight set higher for targets exceeding a specified quantile threshold: $w_i = w$ if $y_i > q_\tau(y)$, and $w_i = 1$ otherwise, where $q_\tau(y)$ is the $\tau$-th quantile of the target distribution and $w > 1$. This approach increases the influence of rare or clinically significant outcomes during optimization, encouraging the model to be more sensitive to underrepresented cases.

During training, only the parameters of the patch embedding, patch-to-token reprogramming module, and output head are updated, while the LLM backbone remains frozen. This modular design enables efficient and robust learning in data-limited neuroimaging settings, leverages the powerful contextual modeling capabilities of pretrained LLMs, and provides a flexible framework for a range of downstream tasks, including regression and classification on dynamic connectome features.

\begin{table*}
\customsmall
\centering
\setlength{\tabcolsep}{2pt}

\begin{subtable}{\textwidth}
\centering
\begin{tabular}{r|ccccccc}
\toprule
\textbf{Diagnosis} &  \texttt{Chebyshev} &   \texttt{Manhattan} & \texttt{Frobenius}&   \texttt{Spectral}  &   \texttt{Nuclear}  &  
    \texttt{Wass-0} &   \texttt{Wass-1}  \\
\midrule
MCI+IMP   & 0.7316\,$\pm$\,0.1439 & 0.7469\,$\pm$\,0.1549 & 0.7296\,$\pm$\,0.1607 & 0.7579\,$\pm$\,0.1509 & 0.7249\,$\pm$\,0.1630 & 0.7700\,$\pm$\,0.1068 & 0.7514\,$\pm$\,0.1511 \\
Normal & 0.5695\,$\pm$\,0.2157 & 0.5708\,$\pm$\,0.2560 & 0.5808\,$\pm$\,0.2116 & 0.5769\,$\pm$\,0.2957 & 0.5684\,$\pm$\,0.1575 & 0.4846\,$\pm$\,0.0870 & 0.5780\,$\pm$\,0.2910 \\
\bottomrule
\end{tabular}
\caption{Llama}
\label{tab:}
\end{subtable}

\vspace{0.4em}

\begin{subtable}{\textwidth}
\centering
\begin{tabular}{r|ccccccc}
\toprule
\textbf{Diagnosis} &  \texttt{Chebyshev} &   \texttt{Manhattan} & \texttt{Frobenius}&   \texttt{Spectral}  &   \texttt{Nuclear}  &  
    \texttt{Wass-0} &   \texttt{Wass-1}  \\
\midrule
MCI+IMP   & 0.7164\,$\pm$\,0.1543 & 0.6864\,$\pm$\,0.1087 & 0.7290\,$\pm$\,0.1660 & 0.7150\,$\pm$\,0.1406 & 0.7415\,$\pm$\,0.1576 & 0.6711\,$\pm$\,0.0813 & 0.6814\,$\pm$\,0.1155 \\
Normal & 0.6022\,$\pm$\,0.1475 & 0.6382\,$\pm$\,0.1027 & 0.5800\,$\pm$\,0.1619 & 0.6099\,$\pm$\,0.1526 & 0.5635\,$\pm$\,0.1480 & 0.6447\,$\pm$\,0.1253 & 0.6409\,$\pm$\,0.1488 \\
\bottomrule
\end{tabular}
\caption{GPT-2}
\label{tab:}
\end{subtable}

\vspace{0.4em}

\begin{subtable}{\textwidth}
\centering
\begin{tabular}{r|ccccccc}
\toprule
\textbf{Diagnosis} &  \texttt{Chebyshev} &   \texttt{Manhattan} & \texttt{Frobenius}&   \texttt{Spectral}  &   \texttt{Nuclear}  &  
    \texttt{Wass-0} &   \texttt{Wass-1}  \\
\midrule
MCI+IMP   & 0.8949\,$\pm$\,0.2164 & 0.8971\,$\pm$\,0.2222 & 0.9034\,$\pm$\,0.2344 & 0.8983\,$\pm$\,0.2210 & 0.8902\,$\pm$\,0.2126 & 0.8955\,$\pm$\,0.2250 & 0.8876\,$\pm$\,0.1889 \\
Normal & 0.3673\,$\pm$\,0.1671 & 0.3696\,$\pm$\,0.1806 & 0.3619\,$\pm$\,0.1864 & 0.3660\,$\pm$\,0.1742 & 0.3752\,$\pm$\,0.1724 & 0.3693\,$\pm$\,0.1806 & 0.3706\,$\pm$\,0.1380 \\
\bottomrule
\end{tabular}
\caption{BERT}
\label{tab:}
\end{subtable}

\caption{MAE ($\pm$ std) for each backbone and connectome distance metric.}
\label{tab:mae_all_train_sizes}
\end{table*}

\section{Early Detection of Alzheimer's}

We evaluate our connectome-informed LLM framework using the OASIS-3 dataset \citep{lamontagne2019oasis}, which provides multimodal neuroimaging and comprehensive clinical assessment data. The sample consists of 677 subjects with resting-state fMRI and clinical diagnostic labels, including 629 with normal cognition, 21 with mild cognitive impairment (MCI), and 27 with cognitive impairment but not meeting MCI criteria (IMP). Diagnostic labels are based on OASIS-3 clinical assessments and enable the identification of early cognitive impairment, which is the primary focus of our study. We exclude subjects with dementia, as early detection and staging are most clinically relevant for timely intervention.

For each subject, BOLD time series are extracted from resting-state fMRI and used to compute dynamic functional connectomes. We quantify dynamic changes in brain connectome topology using a diverse set of well-established distance metrics in connectomics, including \texttt{Chebyshev}, \texttt{Manhattan}, \texttt{Frobenius}, \texttt{spectral}, and \texttt{nuclear} distances, as well as Wasserstein distances for 0-homology (\texttt{Wass-0}) and 1-homology (\texttt{Wass-1}) \citep{edelsbrunner2022computational}. Our primary regression target is the Clinical Dementia Rating Sum of Boxes (\texttt{CDR-SB}), a widely used measure of disease severity and staging in both research and clinical care \citep{albert2011diagnosis}.

The dataset is split by subject into 30\% training, 30\% validation, and 40\% test sets. Given the limited number of MCI and IMP samples, this split ensures that each set contains sufficient examples from each group, which is important for training robust models and reliably evaluating early detection performance. All experiments are repeated across five independent random splits, and results are averaged.

We evaluate three frozen language model backbones: Llama \citep{touvron2023llama}, GPT-2 \citep{radford2019language}, and BERT \citep{devlin2019bert}. Mean absolute error (MAE) is reported separately for the normal cognition group and for the combined impaired group (MCI+IMP). This distinction is clinically meaningful, as early detection of cognitive impairment prior to dementia onset is critical for intervention and prognosis.

Additional details on experiments, data processing, and distance metrics are provided in Appendices~\ref{app:model_config}, \ref{app:dataset}, and \ref{app:distances}.

\subsection{Results and clinical relevance}

Table~\ref{tab:mae_all_train_sizes} presents the MAE and standard deviation for each model backbone and distance metric, evaluated separately for normal cognition and MCI+IMP groups. These results suggest potential for clinical utility when compared to established minimal clinically important difference (MCID) thresholds for \texttt{CDR-SB} in mildly impaired groups, with an MCID set between 1.0 and 2.0 points \citep{muir2024minimal}. Across all model backbones and connectome distance metrics, the MAE for the MCI+IMP group remains well below this threshold, indicating the model's potential for disease monitoring and early detection. Performance is predictably higher (lower MAE) in the normal group and more challenging in the MCI+IMP group, consistent with established knowledge that impaired populations exhibit greater heterogeneity in disease progression and are underrepresented in clinical samples.

Models incorporating global measurements, including Wasserstein and nuclear distances, generally achieve lower MAE compared to those using local measurements such as Manhattan, Frobenius, and Chebyshev distances. This suggests that global properties of the connectome may be more informative for distinguishing early cognitive impairment and tracking disease progression, aligning with clinical understanding that Alzheimer's pathology disrupts brain networks at a global scale.

Early detection of Alzheimer's disease remains a significant challenge, as clinical symptoms are often subtle and overt anatomical changes in the brain may not be detectable until the disease is well advanced. In contrast, functional changes, captured here through dynamic connectome analysis and modeled using LLMs, hold promise for identifying early alterations in brain network organization before irreversible damage occurs. Despite their potential, both functional connectome dynamics and LLM-based modeling remain relatively underexplored as tools for prodromal and preclinical AD detection. Our findings demonstrate that LLMs, with their capacity for transfer learning and generalizability, can extract clinically meaningful patterns from functional brain network data even in data-limited settings that reflect real-world population distributions. Importantly, this is achieved without the need to fine-tune the LLM backbone, as only task-specific parameters are updated, reducing the risk of overfitting. These results highlight the feasibility and promise of leveraging LLMs for sensitive early diagnosis and motivate further exploration of these models and functional network biomarkers for early intervention and monitoring in Alzheimer's disease.

\begin{table}
\customsmall
\centering
\setlength{\tabcolsep}{3.9pt}
\begin{tabular}{r|cccccc}
\toprule
\textbf{Diagnosis} & Default & LL6 & AH4 & TP500 & No RevRIN \\
\midrule
MCI+IMP & 0.6814 & 1.2031 & 1.0461 & 0.6930 & 0.7060 \\
Normal & 0.6409 & 0.7298 & 0.2424 & 0.6311 & 0.5908 \\
\bottomrule
\end{tabular}
\caption{Ablation results for \texttt{Wass-1} MAE. Default: GPT-2, 12 LLM layers (LL), 8 attention heads (AH), 50 text prototypes (TP), with RevRIN.}
\label{tab:ablation}
\end{table}

\paragraph{Ablation study.} Table~\ref{tab:ablation} shows that the default configuration generally yields more favourable MAE for both groups. Adjusting LLM layers (LL6), attention heads (AH4), or text prototypes (TP500) increases error, particularly in MCI+IMP. Excluding RevRIN also reduces performance. Both model architecture and normalization are essential for accurate early detection.
Extensive ablation results are provided in Appendix~\ref{appx:ablation}.

\section*{Limitations}

Despite the strong predictive performance of our connectome-informed LLM framework, interpretability remains a challenge. The process by which dynamic connectome features are translated into clinical predictions by the LLM is not easily understood, which may hinder acceptance in clinical environments that demand clear reasoning for medical decisions. Enhancing transparency represents an important area for future research. Direct involvement of clinicians in evaluating and refining explanation methods could also improve trust and support the practical deployment of such models.

\bibliographystyle{abbrvnat}
\bibliography{reference}

\clearpage
\appendix

\section{Model Configurations}
\label{app:model_config}

We report the default configuration used for connectome-based prediction. The full setup is summarized in Table~\ref{tab:model_config}. We use a frozen language model backbone, either LLaMA (16 layers), GPT-2 (12 layers), or BERT (12 layers). Inputs are segmented into patches of length $L = 8$ with a stride $S = 4$, and projected into a $d_e = 8$ dimensional embedding space. Semantic alignment is performed using $K = 8$ attention heads and $V' = 50$ text prototypes. Models are trained for 30 epochs using the Adam optimizer~\citep{KingmaB14adam} with an initial learning rate of 0.001 and optimized with an order-weighted mean squared error (OW-MSE) loss. This loss assigns higher weight $w = 20$ to samples above the $90^\text{th}$ percentile ($\tau = 0.9$) of the target distribution. Unless otherwise noted, these settings are used throughout our experiments.

\begin{table}[h]
\centering
\setlength{\tabcolsep}{9pt}
\begin{tabular}{l r}
\toprule
\textbf{Configuration} & \textbf{Value} \\
\midrule
LLM Layers: &  \\
\ \ \ Llama & 16 \\
\ \ \ GPT-2 & 12 \\
\ \ \ BERT & 12 \\
Heads $K$ & 8 \\
Text Prototypes $V'$ & 50 \\
Patch Embedding Dim.\ $d_e$ & 8 \\
Patch Length $L$ & 8 \\
Stride $S$ & 4 \\
Initial Learning Rate & $0.001$ \\
Loss & OW-MSE \\
Epochs & 30 \\
\bottomrule
\end{tabular}
\caption{Default model configuration.}
\label{tab:model_config}
\end{table}

\paragraph{Computational resources.} 
The hardware environment consisted of an \textsc{AMD Ryzen Threadripper 7960X} CPU (24 cores), 64\,GB of RAM, a 1\,TB NVMe SSD, and dual \textsc{NVIDIA GeForce RTX 4090} GPUs.

\section{OASIS-3 Dataset}
\label{app:dataset}

This study utilizes the OASIS-3 dataset (Open Access Series of Imaging Studies), a comprehensive, longitudinal resource developed by the Washington University Knight Alzheimer Disease Research Center over a span of 15 years \citep{lamontagne2019oasis}. It was designed to facilitate research into aging and Alzheimer's disease (AD), containing a wide range of neuroimaging and clinical data, including structural MRI, PET scans, and resting-state fMRI data suitable for time series analysis.

To investigate cognitive status, the dataset includes rich clinical assessments based on the Uniform Data Set (UDS) framework, enabling categorization of participants into four groups: normal cognition, mild cognitive impairment (MCI), cognitive impairment not meeting MCI criteria (IMP), and dementia. The classification is guided by variables such as NORMCOG, DEMENTED, and IMPNOMCI. For this work, only participants without a dementia diagnosis were considered, focusing on early-stage cognitive impairment.

The dataset snapshot used was downloaded on December 11, 2024. For inclusion, participants were required to have both T1-weighted (T1w) structural MRI and resting-state fMRI scans. The high-resolution T1w images provide anatomical context for accurate spatial registration of functional data, while resting-state fMRI sequences are used to extract brain activity time series. The data were obtained in BIDS format using scripts from the official OASIS GitHub repository: \url{https://github.com/NrgXnat/oasis-scripts}.

To ensure consistency in diagnosis and adequate temporal coverage, we selected the most recent scan session per participant and filtered sessions by recency and quality. After preprocessing, the final dataset consisted of 677 samples: 629 classified as normal cognition, 21 as MCI, and 27 as IMP. The cohort includes 393 female and 284 male participants, with ages ranging from 42 to 98 years.

\subsection{Preprocessing}

The aim of the preprocessing stage is to minimize noise and imaging artifacts while retaining important neural signals. The first step in this workflow is to transform the raw imaging files into a common, analysis-ready format. Specifically, the original DICOM files are converted to the NIfTI format, and the data are arranged according to the Brain Imaging Data Structure (BIDS) standard. This organization promotes interoperability with a variety of neuroimaging pipelines.

Once the data are formatted and validated in BIDS, we utilize \texttt{fMRIPrep} to carry out consistent preprocessing of the resting-state fMRI scans \citep{esteban2019fmriprep}. \texttt{fMRIPrep} is a widely used preprocessing tool that brings together components from several major neuroimaging software suites, such as FSL, ANTs, and FreeSurfer, to perform a standardized series of steps. The objective is to clean the data and extract key features, including time series information.

Preprocessing with \texttt{fMRIPrep} includes motion correction to address participant movement and slice timing correction to compensate for differences in the timing of slice acquisition. Following these corrections, the processed functional images are aligned with each subject's high-resolution T1w anatomical scan, improving spatial correspondence. The anatomical and functional datasets are then transformed into a standard stereotaxic space (\texttt{MNI152NLin2009cAsym}), enabling comparisons across individuals.

After preprocessing is completed, we extract time series data for different brain regions using the Automated Anatomical Labeling (AAL) atlas, which divides the brain into 116 regions of interest. During this step, we also apply confound regression to remove unwanted sources of variance. This regression model uses 36 parameters, including motion estimates, white matter, and cerebrospinal fluid signals, along with their derivatives and quadratic terms.

\section{Connectome Distance Metrics}
\label{app:distances}

\subsection{Frobenius norm distance}

The Frobenius norm distance provides a quantitative measure of the overall difference between two connectivity matrices. It is defined as the square root of the sum of squared differences between corresponding elements in the matrices.

This approach captures the total geometric deviation across all pairs of regions. It is intuitive, computationally efficient, and particularly appropriate for comparing matrices when a consistent brain atlas is used, ensuring that matrix dimensions and node correspondence are preserved. However, the Frobenius norm is less sensitive to small or localized differences, since the squaring of small values further diminishes their influence on the total distance. This limitation becomes especially relevant in the context of noisy data such as fMRI, where important localized changes may be obscured by large fluctuations driven by noise, which are accentuated by the squaring operation.

\subsection{Manhattan distance}

The Manhattan distance quantifies the overall absolute difference between each corresponding element in two connectivity matrices. Unlike the Frobenius norm, the Manhattan distance does not involve squaring the differences, so it does not disproportionately emphasize large discrepancies. As a result, this metric can be more attuned to subtle or localized changes in connectivity, and is less likely to be dominated by large outliers, which might result from noise. However, this property also means that the Manhattan distance might be more influenced by minor, random variations.

\subsection{Chebyshev distance}

The Chebyshev distance measures the greatest absolute difference between any pair of corresponding elements in two vectors or matrices. This metric is particularly sensitive to the largest individual deviation within the set, as it completely disregards the size of all other differences.

By concentrating solely on the maximum deviation, the Chebyshev distance becomes very responsive to outlier values. Its value depends entirely on the most pronounced change, regardless of the overall consistency or variation among the other elements. This characteristic makes it well-suited for situations where extreme values carry significant meaning for the analysis. However, this property also means that the metric can be easily influenced by noise or measurement errors, as these can disproportionately affect the result.

\subsection{Nuclear norm distance}

The nuclear norm distance offers a measure of similarity between two matrices by considering the sum of their singular values after taking their difference. Instead of focusing on specific entries, this metric highlights differences in the broader structural patterns or dominant features within the matrices. It is particularly effective for capturing changes in low-rank structures, which often summarize the most significant information in the data. On the other hand, the nuclear norm distance may be less responsive to minor, localized variations.

\subsection{Spectral distance}

The spectral distance offers a way to compare two matrices by examining the differences in their eigenvalue spectra. Instead of focusing on individual elements, this measure captures how the overall structure of the connectivity graph changes. By considering the entire set of eigenvalues for each matrix, spectral distance highlights global features of the network, such as patterns of connectivity, network stability, and community structure.

This metric is especially useful for identifying substantial shifts in the network's organization, including changes involving hub nodes or the emergence and dissolution of communities. Because spectral distance reflects the broader architecture rather than localized changes, it is well suited for detecting significant reconfigurations in complex systems like brain networks, even when the specific details of individual connections are different.

\subsection{Wasserstein distance for 0-homology and 1-homology}

Persistent graph homology offers a powerful framework to analyze, describe, and measure structural patterns in human connectomes \citep{songdechakraiwut2021topological,songdechakraiwut2023topological}. This approach leverages topological invariants of a graph, notably connected components (captured by the 0th homology group) and cycles (represented by the 1st homology group, also called the cycle rank). 

Given a connectivity matrix $C$, we generate a series of binary graphs $G_\tau$ by applying a threshold $\tau$ to the edge weights. For each $\tau$, an edge exists in $G_\tau$ if its weight is at least $\tau$, and is absent otherwise. As $\tau$ increases, the graph progressively loses edges, creating a sequence of graphs known as a filtration:

\[
G_{\tau_1} \supseteq G_{\tau_2} \supseteq \cdots \supseteq G_{\tau_k}
\]

Persistent homology tracks how the topological features of the graph, such as connected components and cycles, emerge (birth) and vanish (death) throughout this filtration process. Each feature's birth and death is recorded as a point $(b, d)$ in a persistence diagram, where $b$ and $d$ are the respective filtration values.

As $\tau$ increases, the graph becomes sparser, so the number of connected components increases monotonically, while the number of cycles decreases. To capture these topological changes, it is sufficient to keep track of the birth filtration values of all connected components, denoted by the set $\mathcal{B}_0$, and the death filtration values of all cycles, denoted by $\mathcal{D}_1$. Both of these sets have cardinality equal to the number of brain regions $n$. Together, these sets provide a succinct summary of the essential topological features that persist through the filtration.

To quantitatively compare persistence diagrams, the Wasserstein distance is widely used. This metric is especially important because of its stability properties \citep{skraba2020wasserstein}. Given two connectivity matrices $C$ and $C'$, the $p$-Wasserstein distance for 0-homology is defined as
\[
W_p^{(0)}(\mathcal{B}_0, \mathcal{B}_0') = \left( \sum_{i=1}^n |b_i - b'_i|^p \right)^{1/p}
\]
while for 1-homology, it is defined as
\[
W_p^{(1)}(\mathcal{D}_1, \mathcal{D}_1') = \left( \sum_{i=1}^n |d_i - d'_i|^p \right)^{1/p},
\]
where the birth values $b_i$ and $b'_i$, as well as the death values $d_i$ and $d'_i$, are optimally matched in sorted order between the two diagrams~\citep{songdechakraiwut2023wasserstein}. Since all the persistence diagrams are obtained from graphs with the same number of brain regions, the sets of birth and death values always have the same size, ensuring that these distances are well-defined.

\section{Detailed Ablation Results}
\label{appx:ablation}

The full ablation results are presented in Tables~\ref{tab:mae_weight}, \ref{tab:mae_all_llmlayers_ablation}, \ref{tab:mae_all_nheads_ablation}, \ref{tab:mae_all_prototypes_ablation}, and~\ref{tab:mae_all_instancenorm_robust_ablation}.

\begin{table*}
\customsmall
\centering
\setlength{\tabcolsep}{2pt}

\begin{subtable}{\textwidth}
\centering
\begin{tabular}{r|ccccccc}
\toprule
\textbf{Diagnosis} &  \texttt{Chebyshev} &   \texttt{Manhattan} & \texttt{Frobenius}&   \texttt{Spectral}  &   \texttt{Nuclear}  &  
    \texttt{Wass-0} &   \texttt{Wass-1}  \\
\midrule
MCI+IMP   & 0.8611\,$\pm$\,0.1409 & 0.8008\,$\pm$\,0.1759 & 0.8662\,$\pm$\,0.1445 & 0.7863\,$\pm$\,0.1487 & 0.7973\,$\pm$\,0.1805 & 0.8400\,$\pm$\,0.1470 & 0.8153\,$\pm$\,0.2176 \\
Normal & 0.4051\,$\pm$\,0.1242 & 0.4754\,$\pm$\,0.1373 & 0.3985\,$\pm$\,0.1484 & 0.4844\,$\pm$\,0.1125 & 0.4862\,$\pm$\,0.1453 & 0.4220\,$\pm$\,0.1179 & 0.4603\,$\pm$\,0.1805 \\
\bottomrule
\end{tabular}
\caption{Llama}
\label{tab:}
\end{subtable}

\vspace{0.5em}

\begin{subtable}{\textwidth}
\centering
\begin{tabular}{r|ccccccc}
\toprule
\textbf{Diagnosis} &  \texttt{Chebyshev} &   \texttt{Manhattan} & \texttt{Frobenius}&   \texttt{Spectral}  &   \texttt{Nuclear}  &  
    \texttt{Wass-0} &   \texttt{Wass-1}  \\
\midrule
MCI+IMP   & 0.7573\,$\pm$\,0.0794 & 0.7533\,$\pm$\,0.0913 & 0.7814\,$\pm$\,0.1208 & 0.7313\,$\pm$\,0.0874 & 0.8372\,$\pm$\,0.1681 & 0.7066\,$\pm$\,0.0918 & 0.7574\,$\pm$\,0.0682 \\
Normal & 0.5059\,$\pm$\,0.0626 & 0.5176\,$\pm$\,0.0776 & 0.4945\,$\pm$\,0.1169 & 0.5674\,$\pm$\,0.1202 & 0.4141\,$\pm$\,0.1288 & 0.6547\,$\pm$\,0.3011 & 0.5015\,$\pm$\,0.0731 \\
\bottomrule
\end{tabular}
\caption{GPT-2}
\label{tab:}
\end{subtable}

\vspace{0.5em}

\begin{subtable}{\textwidth}
\centering
\begin{tabular}{r|ccccccc}
\toprule
\textbf{Diagnosis} &  \texttt{Chebyshev} &   \texttt{Manhattan} & \texttt{Frobenius}&   \texttt{Spectral}  &   \texttt{Nuclear}  &  
    \texttt{Wass-0} &   \texttt{Wass-1}  \\
\midrule
MCI+IMP   & 1.0027\,$\pm$\,0.1776 & 0.9957\,$\pm$\,0.1940 & 0.9962\,$\pm$\,0.1848 & 0.9884\,$\pm$\,0.1979 & 0.9969\,$\pm$\,0.1969 & 1.0020\,$\pm$\,0.1733 & 0.9911\,$\pm$\,0.1839 \\
Normal & 0.2607\,$\pm$\,0.1061 & 0.2660\,$\pm$\,0.1170 & 0.2674\,$\pm$\,0.1128 & 0.2735\,$\pm$\,0.1248 & 0.2645\,$\pm$\,0.1180 & 0.2600\,$\pm$\,0.1030 & 0.2705\,$\pm$\,0.1112 \\
\bottomrule
\end{tabular}
\caption{BERT}
\label{tab:}
\end{subtable}

\caption{MAE ($\pm$ std) with OW-MSE ($w{=}10$, $\tau{=}0.9$) across models and groups.}
\label{tab:mae_weight}
\end{table*}

\begin{table*}
\customsmall
\centering
\setlength{\tabcolsep}{1.5pt}

\begin{subtable}{\textwidth}
\centering
\begin{tabular}{l r|ccccccc}
\toprule
Layers & \textbf{Diagnosis} &  \texttt{Chebyshev} &   \texttt{Manhattan} & \texttt{Frobenius} &   \texttt{Spectral}  &   \texttt{Nuclear}  &  \texttt{Wass-0} &   \texttt{Wass-1}  \\
\midrule
\multirow{2}{*}{8} 
    & MCI+IMP & 0.8628\,$\pm$\,0.2122 & 0.8564\,$\pm$\,0.2270 & 0.8451\,$\pm$\,0.2351 & 0.8610\,$\pm$\,0.2340 & 0.8647\,$\pm$\,0.2376 & 0.8696\,$\pm$\,0.2247 & 0.8697\,$\pm$\,0.2273 \\
    & Normal  & 0.4105\,$\pm$\,0.1637 & 0.4434\,$\pm$\,0.2105 & 0.4578\,$\pm$\,0.2359 & 0.4397\,$\pm$\,0.2295 & 0.4244\,$\pm$\,0.1965 & 0.4321\,$\pm$\,0.2200 & 0.4118\,$\pm$\,0.1830 \\
\multirow{2}{*}{16} 
    & MCI+IMP & 0.7316\,$\pm$\,0.1439 & 0.7469\,$\pm$\,0.1549 & 0.7296\,$\pm$\,0.1607 & 0.7579\,$\pm$\,0.1509 & 0.7249\,$\pm$\,0.1630 & 0.7700\,$\pm$\,0.1068 & 0.7514\,$\pm$\,0.1511 \\
    & Normal  & 0.5695\,$\pm$\,0.2157 & 0.5708\,$\pm$\,0.2560 & 0.5808\,$\pm$\,0.2116 & 0.5769\,$\pm$\,0.2957 & 0.5684\,$\pm$\,0.1575 & 0.4846\,$\pm$\,0.0870 & 0.5780\,$\pm$\,0.2910 \\
\bottomrule
\end{tabular}
\caption{Llama}
\label{tab:mae_llama_llmlayers}
\end{subtable}

\vspace{0.5em}

\begin{subtable}{\textwidth}
\centering
\begin{tabular}{l r|ccccccc}
\toprule
Layers & \textbf{Diagnosis} &  \texttt{Chebyshev} &   \texttt{Manhattan} & \texttt{Frobenius} &   \texttt{Spectral}  &   \texttt{Nuclear}  &  \texttt{Wass-0} &   \texttt{Wass-1}  \\
\midrule
\multirow{2}{*}{6} 
    & MCI+IMP & 1.0092\,$\pm$\,0.4147 & 1.0837\,$\pm$\,0.4219 & 0.8142\,$\pm$\,0.2048 & 1.0321\,$\pm$\,0.4195 & 1.0179\,$\pm$\,0.3568 & 1.1969\,$\pm$\,0.4834 & 1.2031\,$\pm$\,0.5392 \\
    & Normal  & 0.6975\,$\pm$\,0.5616 & 0.6408\,$\pm$\,0.6015 & 0.6237\,$\pm$\,0.5159 & 0.5663\,$\pm$\,0.5275 & 0.6851\,$\pm$\,0.5605 & 0.4433\,$\pm$\,0.1393 & 0.7298\,$\pm$\,0.5852 \\
\multirow{2}{*}{12} 
    & MCI+IMP & 0.7164\,$\pm$\,0.1543 & 0.6864\,$\pm$\,0.1087 & 0.7290\,$\pm$\,0.1660 & 0.7150\,$\pm$\,0.1406 & 0.7415\,$\pm$\,0.1576 & 0.6711\,$\pm$\,0.0813 & 0.6814\,$\pm$\,0.1155 \\
    & Normal  & 0.6022\,$\pm$\,0.1475 & 0.6382\,$\pm$\,0.1027 & 0.5800\,$\pm$\,0.1619 & 0.6099\,$\pm$\,0.1526 & 0.5635\,$\pm$\,0.1480 & 0.6447\,$\pm$\,0.1253 & 0.6409\,$\pm$\,0.1488 \\
\bottomrule
\end{tabular}
\caption{GPT-2}
\label{tab:mae_gpt2_llmlayers}
\end{subtable}

\vspace{0.5em}

\begin{subtable}{\textwidth}
\centering
\begin{tabular}{l r|ccccccc}
\toprule
Layers & \textbf{Diagnosis} &  \texttt{Chebyshev} &   \texttt{Manhattan} & \texttt{Frobenius} &   \texttt{Spectral}  &   \texttt{Nuclear}  &  \texttt{Wass-0} &   \texttt{Wass-1}  \\
\midrule
\multirow{2}{*}{6} 
    & MCI+IMP & 0.8554\,$\pm$\,0.1860 & 0.8540\,$\pm$\,0.1703 & 0.8431\,$\pm$\,0.1876 & 0.8377\,$\pm$\,0.1929 & 0.8411\,$\pm$\,0.1943 & 0.8444\,$\pm$\,0.1825 & 0.8454\,$\pm$\,0.1863 \\
    & Normal  & 0.4386\,$\pm$\,0.2039 & 0.4187\,$\pm$\,0.1506 & 0.4501\,$\pm$\,0.2028 & 0.4543\,$\pm$\,0.2021 & 0.4522\,$\pm$\,0.2063 & 0.4480\,$\pm$\,0.2010 & 0.4496\,$\pm$\,0.2032 \\
\multirow{2}{*}{12} 
    & MCI+IMP & 0.8949\,$\pm$\,0.2164 & 0.8971\,$\pm$\,0.2222 & 0.9034\,$\pm$\,0.2344 & 0.8983\,$\pm$\,0.2210 & 0.8902\,$\pm$\,0.2126 & 0.8955\,$\pm$\,0.2250 & 0.8876\,$\pm$\,0.1889 \\
    & Normal  & 0.3673\,$\pm$\,0.1671 & 0.3696\,$\pm$\,0.1806 & 0.3619\,$\pm$\,0.1864 & 0.3660\,$\pm$\,0.1742 & 0.3752\,$\pm$\,0.1724 & 0.3693\,$\pm$\,0.1806 & 0.3706\,$\pm$\,0.1380 \\
\bottomrule
\end{tabular}
\caption{BERT}
\label{tab:mae_bert_llmlayers}
\end{subtable}

\caption{MAE ($\pm$ std) for ablations by number of LLM layers and model.}
\label{tab:mae_all_llmlayers_ablation}
\end{table*}

\begin{table*}
\customsmall
\centering
\setlength{\tabcolsep}{2.2pt}

\begin{subtable}{\textwidth}
\centering
\begin{tabular}{l r|ccccccc}
\toprule
$K$ & \textbf{Diagnosis} &  \texttt{Chebyshev} &   \texttt{Manhattan} & \texttt{Frobenius} &   \texttt{Spectral}  &   \texttt{Nuclear}  &  \texttt{Wass-0} &   \texttt{Wass-1}  \\
\midrule
\multirow{2}{*}{2} 
    & MCI+IMP & 0.9272\,$\pm$\,0.2409 & 0.9218\,$\pm$\,0.2478 & 0.9114\,$\pm$\,0.2381 & 0.8910\,$\pm$\,0.2836 & 0.9037\,$\pm$\,0.2421 & 0.9175\,$\pm$\,0.2528 & 0.9016\,$\pm$\,0.2572 \\
    & Normal  & 0.3473\,$\pm$\,0.2080 & 0.3509\,$\pm$\,0.2181 & 0.3617\,$\pm$\,0.2145 & 0.3914\,$\pm$\,0.2398 & 0.3703\,$\pm$\,0.2113 & 0.3618\,$\pm$\,0.2237 & 0.3676\,$\pm$\,0.2138 \\
\multirow{2}{*}{4} 
    & MCI+IMP & 0.7991\,$\pm$\,0.2174 & 0.8092\,$\pm$\,0.1828 & 0.8429\,$\pm$\,0.2076 & 0.7618\,$\pm$\,0.2315 & 0.7525\,$\pm$\,0.2349 & 0.7886\,$\pm$\,0.2086 & 0.7769\,$\pm$\,0.2283 \\
    & Normal  & 0.4892\,$\pm$\,0.2560 & 0.4628\,$\pm$\,0.1445 & 0.4132\,$\pm$\,0.1378 & 0.5518\,$\pm$\,0.2948 & 0.5873\,$\pm$\,0.3405 & 0.4806\,$\pm$\,0.1610 & 0.5415\,$\pm$\,0.3174 \\
\multirow{2}{*}{6} 
    & MCI+IMP & 0.8383\,$\pm$\,0.1611 & 0.8356\,$\pm$\,0.1569 & 0.8314\,$\pm$\,0.1600 & 0.8311\,$\pm$\,0.1597 & 0.8451\,$\pm$\,0.1560 & 0.8435\,$\pm$\,0.1620 & 0.8300\,$\pm$\,0.1623 \\
    & Normal  & 0.4802\,$\pm$\,0.2712 & 0.4725\,$\pm$\,0.2301 & 0.4926\,$\pm$\,0.2698 & 0.4741\,$\pm$\,0.2269 & 0.4617\,$\pm$\,0.2105 & 0.4634\,$\pm$\,0.2253 & 0.4796\,$\pm$\,0.2636 \\
\multirow{2}{*}{8} 
    & MCI+IMP & 0.7316\,$\pm$\,0.1439 & 0.7469\,$\pm$\,0.1549 & 0.7296\,$\pm$\,0.1607 & 0.7579\,$\pm$\,0.1509 & 0.7249\,$\pm$\,0.1630 & 0.7700\,$\pm$\,0.1068 & 0.7514\,$\pm$\,0.1511 \\
    & Normal  & 0.5695\,$\pm$\,0.2157 & 0.5708\,$\pm$\,0.2560 & 0.5808\,$\pm$\,0.2116 & 0.5769\,$\pm$\,0.2957 & 0.5684\,$\pm$\,0.1575 & 0.4846\,$\pm$\,0.0870 & 0.5780\,$\pm$\,0.2910 \\
\bottomrule
\end{tabular}
\caption{Llama}
\label{tab:mae_llama_nheads}
\end{subtable}

\vspace{0.5em}

\begin{subtable}{\textwidth}
\centering
\begin{tabular}{l r|ccccccc}
\toprule
$K$ & \textbf{Diagnosis} &  \texttt{Chebyshev} &   \texttt{Manhattan} & \texttt{Frobenius} &   \texttt{Spectral}  &   \texttt{Nuclear}  &  \texttt{Wass-0} &   \texttt{Wass-1}  \\
\midrule
\multirow{2}{*}{2} 
    & MCI+IMP & 1.0210\,$\pm$\,0.4201 & 0.9382\,$\pm$\,0.2548 & 0.9658\,$\pm$\,0.2246 & 0.9224\,$\pm$\,0.2990 & 0.9119\,$\pm$\,0.2336 & 0.9609\,$\pm$\,0.3230 & 0.8829\,$\pm$\,0.1348 \\
    & Normal  & 0.4890\,$\pm$\,0.2033 & 0.4075\,$\pm$\,0.2292 & 0.3614\,$\pm$\,0.1731 & 0.4785\,$\pm$\,0.3084 & 0.4114\,$\pm$\,0.2503 & 0.4568\,$\pm$\,0.2044 & 0.4134\,$\pm$\,0.2812 \\
\multirow{2}{*}{4} 
    & MCI+IMP & 0.9176\,$\pm$\,0.2731 & 0.8841\,$\pm$\,0.2082 & 0.9617\,$\pm$\,0.1815 & 0.9469\,$\pm$\,0.1692 & 0.9642\,$\pm$\,0.2632 & 0.9910\,$\pm$\,0.1671 & 1.0461\,$\pm$\,0.2406 \\
    & Normal  & 0.3754\,$\pm$\,0.2670 & 0.3986\,$\pm$\,0.2307 & 0.3184\,$\pm$\,0.2400 & 0.3250\,$\pm$\,0.1882 & 0.3270\,$\pm$\,0.2546 & 0.2747\,$\pm$\,0.1819 & 0.2424\,$\pm$\,0.2287 \\
\multirow{2}{*}{6} 
    & MCI+IMP & 0.8971\,$\pm$\,0.2008 & 0.8503\,$\pm$\,0.2344 & 0.9014\,$\pm$\,0.0744 & 0.8495\,$\pm$\,0.1281 & 0.9036\,$\pm$\,0.2074 & 0.9590\,$\pm$\,0.1026 & 0.9369\,$\pm$\,0.2088 \\
    & Normal  & 0.3614\,$\pm$\,0.1668 & 0.4359\,$\pm$\,0.2001 & 0.3539\,$\pm$\,0.0392 & 0.3963\,$\pm$\,0.0589 & 0.3517\,$\pm$\,0.1563 & 0.3025\,$\pm$\,0.0816 & 0.3222\,$\pm$\,0.1476 \\
\multirow{2}{*}{8} 
    & MCI+IMP & 0.7164\,$\pm$\,0.1543 & 0.6864\,$\pm$\,0.1087 & 0.7290\,$\pm$\,0.1660 & 0.7150\,$\pm$\,0.1406 & 0.7415\,$\pm$\,0.1576 & 0.6711\,$\pm$\,0.0813 & 0.6814\,$\pm$\,0.1155 \\
    & Normal  & 0.6022\,$\pm$\,0.1475 & 0.6382\,$\pm$\,0.1027 & 0.5800\,$\pm$\,0.1619 & 0.6099\,$\pm$\,0.1526 & 0.5635\,$\pm$\,0.1480 & 0.6447\,$\pm$\,0.1253 & 0.6409\,$\pm$\,0.1488 \\
\bottomrule
\end{tabular}
\caption{GPT-2}
\label{tab:mae_gpt2_nheads}
\end{subtable}

\vspace{0.5em}

\begin{subtable}{\textwidth}
\centering
\begin{tabular}{l r|ccccccc}
\toprule
$K$ & \textbf{Diagnosis} &  \texttt{Chebyshev} &   \texttt{Manhattan} & \texttt{Frobenius} &   \texttt{Spectral}  &   \texttt{Nuclear}  &  \texttt{Wass-0} &   \texttt{Wass-1}  \\
\midrule
\multirow{2}{*}{2} 
    & MCI+IMP & 1.0002\,$\pm$\,0.2213 & 0.9990\,$\pm$\,0.2227 & 0.9965\,$\pm$\,0.2192 & 0.9979\,$\pm$\,0.2218 & 0.9916\,$\pm$\,0.2197 & 0.9908\,$\pm$\,0.2187 & 0.9917\,$\pm$\,0.2233 \\
    & Normal  & 0.2805\,$\pm$\,0.2021 & 0.2790\,$\pm$\,0.1990 & 0.2802\,$\pm$\,0.1946 & 0.2808\,$\pm$\,0.1975 & 0.2869\,$\pm$\,0.1986 & 0.2867\,$\pm$\,0.1961 & 0.2882\,$\pm$\,0.2049 \\
\multirow{2}{*}{4} 
    & MCI+IMP & 0.8383\,$\pm$\,0.0707 & 0.8434\,$\pm$\,0.0659 & 0.8393\,$\pm$\,0.0703 & 0.8292\,$\pm$\,0.0659 & 0.8378\,$\pm$\,0.0691 & 0.8436\,$\pm$\,0.0753 & 0.8404\,$\pm$\,0.0704 \\
    & Normal  & 0.4075\,$\pm$\,0.0626 & 0.4031\,$\pm$\,0.0613 & 0.4105\,$\pm$\,0.0717 & 0.4179\,$\pm$\,0.0687 & 0.4111\,$\pm$\,0.0705 & 0.4037\,$\pm$\,0.0558 & 0.4084\,$\pm$\,0.0699 \\
\multirow{2}{*}{6} 
    & MCI+IMP & 0.8840\,$\pm$\,0.2156 & 0.8855\,$\pm$\,0.2189 & 0.8880\,$\pm$\,0.2216 & 0.8933\,$\pm$\,0.2219 & 0.8886\,$\pm$\,0.2201 & 0.8838\,$\pm$\,0.2185 & 0.8860\,$\pm$\,0.2202 \\
    & Normal  & 0.4238\,$\pm$\,0.2589 & 0.4207\,$\pm$\,0.2582 & 0.4183\,$\pm$\,0.2579 & 0.4118\,$\pm$\,0.2583 & 0.4148\,$\pm$\,0.2527 & 0.4203\,$\pm$\,0.2512 & 0.4197\,$\pm$\,0.2551 \\
\multirow{2}{*}{8} 
    & MCI+IMP & 0.8949\,$\pm$\,0.2164 & 0.8971\,$\pm$\,0.2222 & 0.9034\,$\pm$\,0.2344 & 0.8983\,$\pm$\,0.2210 & 0.8902\,$\pm$\,0.2126 & 0.8955\,$\pm$\,0.2250 & 0.8876\,$\pm$\,0.1889 \\
    & Normal  & 0.3673\,$\pm$\,0.1671 & 0.3696\,$\pm$\,0.1806 & 0.3619\,$\pm$\,0.1864 & 0.3660\,$\pm$\,0.1742 & 0.3752\,$\pm$\,0.1724 & 0.3693\,$\pm$\,0.1806 & 0.3706\,$\pm$\,0.1380 \\
\bottomrule
\end{tabular}
\caption{BERT}
\label{tab:mae_bert_nheads}
\end{subtable}

\caption{MAE ($\pm$ std) for ablations by number of attention heads and model.}
\label{tab:mae_all_nheads_ablation}
\end{table*}

\begin{table*}
\customsmall
\centering
\setlength{\tabcolsep}{2.2pt}

\begin{subtable}{\textwidth}
\centering
\begin{tabular}{l r|ccccccc}
\toprule
$V'$ & \textbf{Diagnosis} &  \texttt{Chebyshev} &   \texttt{Manhattan} & \texttt{Frobenius} &   \texttt{Spectral}  &   \texttt{Nuclear}  &  \texttt{Wass-0} &   \texttt{Wass-1}  \\
\midrule
\multirow{2}{*}{50} 
    & MCI+IMP & 0.7316\,$\pm$\,0.1439 & 0.7469\,$\pm$\,0.1549 & 0.7296\,$\pm$\,0.1607 & 0.7579\,$\pm$\,0.1509 & 0.7249\,$\pm$\,0.1630 & 0.7700\,$\pm$\,0.1068 & 0.7514\,$\pm$\,0.1511 \\
    & Normal  & 0.5695\,$\pm$\,0.2157 & 0.5708\,$\pm$\,0.2560 & 0.5808\,$\pm$\,0.2116 & 0.5769\,$\pm$\,0.2957 & 0.5684\,$\pm$\,0.1575 & 0.4846\,$\pm$\,0.0870 & 0.5780\,$\pm$\,0.2910 \\
\multirow{2}{*}{100} 
    & MCI+IMP & 0.8003\,$\pm$\,0.2578 & 0.7517\,$\pm$\,0.2107 & 0.7360\,$\pm$\,0.2128 & 0.7527\,$\pm$\,0.1977 & 0.7972\,$\pm$\,0.2098 & 0.7821\,$\pm$\,0.2145 & 0.8080\,$\pm$\,0.2234 \\
    & Normal  & 0.5129\,$\pm$\,0.2526 & 0.5391\,$\pm$\,0.1930 & 0.5980\,$\pm$\,0.2314 & 0.5609\,$\pm$\,0.2032 & 0.4780\,$\pm$\,0.1998 & 0.4850\,$\pm$\,0.2042 & 0.4647\,$\pm$\,0.2117 \\
\multirow{2}{*}{500} 
    & MCI+IMP & 0.7237\,$\pm$\,0.1498 & 0.7427\,$\pm$\,0.1840 & 0.7499\,$\pm$\,0.1997 & 0.7416\,$\pm$\,0.1547 & 0.6978\,$\pm$\,0.1166 & 0.7472\,$\pm$\,0.2088 & 0.7655\,$\pm$\,0.2030 \\
    & Normal  & 0.5919\,$\pm$\,0.1420 & 0.5752\,$\pm$\,0.1833 & 0.5455\,$\pm$\,0.1906 & 0.5585\,$\pm$\,0.1506 & 0.6263\,$\pm$\,0.1116 & 0.5812\,$\pm$\,0.2310 & 0.5394\,$\pm$\,0.1932 \\
\bottomrule
\end{tabular}
\caption{Llama}
\label{tab:mae_llama_prototypes}
\end{subtable}

\vspace{0.5em}

\begin{subtable}{\textwidth}
\centering
\begin{tabular}{l r|ccccccc}
\toprule
$V'$ & \textbf{Diagnosis} &  \texttt{Chebyshev} &   \texttt{Manhattan} & \texttt{Frobenius} &   \texttt{Spectral}  &   \texttt{Nuclear}  &  \texttt{Wass-0} &   \texttt{Wass-1}  \\
\midrule
\multirow{2}{*}{50} 
    & MCI+IMP & 0.7164\,$\pm$\,0.1543 & 0.6864\,$\pm$\,0.1087 & 0.7290\,$\pm$\,0.1660 & 0.7150\,$\pm$\,0.1406 & 0.7415\,$\pm$\,0.1576 & 0.6711\,$\pm$\,0.0813 & 0.6814\,$\pm$\,0.1155 \\
    & Normal  & 0.6022\,$\pm$\,0.1475 & 0.6382\,$\pm$\,0.1027 & 0.5800\,$\pm$\,0.1619 & 0.6099\,$\pm$\,0.1526 & 0.5635\,$\pm$\,0.1480 & 0.6447\,$\pm$\,0.1253 & 0.6409\,$\pm$\,0.1488 \\
\multirow{2}{*}{100} 
    & MCI+IMP & 1.0054\,$\pm$\,0.3277 & 0.8352\,$\pm$\,0.1379 & 1.0922\,$\pm$\,0.4965 & 0.8896\,$\pm$\,0.1802 & 1.0778\,$\pm$\,0.5393 & 0.9287\,$\pm$\,0.3745 & 0.8219\,$\pm$\,0.1868 \\
    & Normal  & 0.3362\,$\pm$\,0.1383 & 0.4468\,$\pm$\,0.1633 & 0.4399\,$\pm$\,0.2334 & 0.3979\,$\pm$\,0.1962 & 0.4325\,$\pm$\,0.2506 & 0.4377\,$\pm$\,0.1782 & 0.5067\,$\pm$\,0.3353 \\
\multirow{2}{*}{500} 
    & MCI+IMP & 0.7384\,$\pm$\,0.2113 & 0.8288\,$\pm$\,0.2491 & 0.7894\,$\pm$\,0.1919 & 0.7604\,$\pm$\,0.1847 & 0.7321\,$\pm$\,0.2318 & 0.7823\,$\pm$\,0.2055 & 0.6930\,$\pm$\,0.1907 \\
    & Normal  & 0.5674\,$\pm$\,0.2027 & 0.4555\,$\pm$\,0.2055 & 0.4807\,$\pm$\,0.1415 & 0.5366\,$\pm$\,0.1584 & 0.5868\,$\pm$\,0.2742 & 0.4988\,$\pm$\,0.1754 & 0.6311\,$\pm$\,0.2094 \\
\bottomrule
\end{tabular}
\caption{GPT-2}
\label{tab:mae_gpt2_prototypes}
\end{subtable}

\vspace{0.5em}

\begin{subtable}{\textwidth}
\centering
\begin{tabular}{l r|ccccccc}
\toprule
$V'$ & \textbf{Diagnosis} &  \texttt{Chebyshev} &   \texttt{Manhattan} & \texttt{Frobenius} &   \texttt{Spectral}  &   \texttt{Nuclear}  &  \texttt{Wass-0} &   \texttt{Wass-1}  \\
\midrule
\multirow{2}{*}{50} 
    & MCI+IMP & 0.8949\,$\pm$\,0.2164 & 0.8971\,$\pm$\,0.2222 & 0.9034\,$\pm$\,0.2344 & 0.8983\,$\pm$\,0.2210 & 0.8902\,$\pm$\,0.2126 & 0.8955\,$\pm$\,0.2250 & 0.8876\,$\pm$\,0.1889 \\
    & Normal  & 0.3673\,$\pm$\,0.1671 & 0.3696\,$\pm$\,0.1806 & 0.3619\,$\pm$\,0.1864 & 0.3660\,$\pm$\,0.1742 & 0.3752\,$\pm$\,0.1724 & 0.3693\,$\pm$\,0.1806 & 0.3706\,$\pm$\,0.1380 \\
\multirow{2}{*}{100} 
    & MCI+IMP & 0.8496\,$\pm$\,0.2196 & 0.8435\,$\pm$\,0.2159 & 0.9355\,$\pm$\,0.1641 & 0.8649\,$\pm$\,0.2033 & 0.9549\,$\pm$\,0.1812 & 0.8506\,$\pm$\,0.2218 & 0.8474\,$\pm$\,0.2095 \\
    & Normal  & 0.4151\,$\pm$\,0.1745 & 0.4208\,$\pm$\,0.1728 & 0.3184\,$\pm$\,0.1331 & 0.3899\,$\pm$\,0.1441 & 0.3027\,$\pm$\,0.1551 & 0.4137\,$\pm$\,0.1778 & 0.4156\,$\pm$\,0.1622 \\
\multirow{2}{*}{500} 
    & MCI+IMP & 0.8551\,$\pm$\,0.1880 & 0.8838\,$\pm$\,0.1812 & 0.8646\,$\pm$\,0.1796 & 0.8687\,$\pm$\,0.1976 & 0.8678\,$\pm$\,0.1870 & 0.8438\,$\pm$\,0.1929 & 0.8816\,$\pm$\,0.1846 \\
    & Normal  & 0.4077\,$\pm$\,0.1400 & 0.3789\,$\pm$\,0.1329 & 0.3999\,$\pm$\,0.1375 & 0.3962\,$\pm$\,0.1482 & 0.3950\,$\pm$\,0.1357 & 0.4233\,$\pm$\,0.1477 & 0.3844\,$\pm$\,0.1374 \\
\bottomrule
\end{tabular}
\caption{BERT}
\label{tab:mae_bert_prototypes}
\end{subtable}

\caption{MAE ($\pm$ std) for ablations by number of text prototypes and model.}
\label{tab:mae_all_prototypes_ablation}
\end{table*}

\begin{table*}
\customsmall
\centering
\setlength{\tabcolsep}{1.5pt}

\begin{subtable}{\textwidth}
\centering
\begin{tabular}{l r|ccccccc}
\toprule
RevRIN & \textbf{Diagnosis} &  \texttt{Chebyshev} &   \texttt{Manhattan} & \texttt{Frobenius} &   \texttt{Spectral}  &   \texttt{Nuclear}  &  \texttt{Wass-0} &   \texttt{Wass-1}  \\
\midrule
\multirow{2}{*}{No} 
    & MCI+IMP & 0.7864\,$\pm$\,0.1396 & 0.7689\,$\pm$\,0.1520 & 0.7073\,$\pm$\,0.1513 & 0.7010\,$\pm$\,0.1684 & 0.7094\,$\pm$\,0.2299 & 0.7134\,$\pm$\,0.1665 & 0.7517\,$\pm$\,0.1337 \\
    & Normal  & 0.5599\,$\pm$\,0.3304 & 0.4669\,$\pm$\,0.1408 & 0.6110\,$\pm$\,0.2421 & 0.5308\,$\pm$\,0.2153 & 0.4531\,$\pm$\,0.0993 & 0.6097\,$\pm$\,0.1755 & 0.5780\,$\pm$\,0.2910 \\
\multirow{2}{*}{Yes} 
    & MCI+IMP & 0.7316\,$\pm$\,0.1439 & 0.7469\,$\pm$\,0.1549 & 0.7296\,$\pm$\,0.1607 & 0.7579\,$\pm$\,0.1509 & 0.7249\,$\pm$\,0.1630 & 0.7700\,$\pm$\,0.1068 & 0.7514\,$\pm$\,0.1511 \\
    & Normal  & 0.5695\,$\pm$\,0.2157 & 0.5708\,$\pm$\,0.2560 & 0.5808\,$\pm$\,0.2116 & 0.5769\,$\pm$\,0.2957 & 0.5684\,$\pm$\,0.1575 & 0.4846\,$\pm$\,0.0870 & 0.5780\,$\pm$\,0.2910 \\
\bottomrule
\end{tabular}
\caption{LLAMA}
\label{tab:mae_llama_instancenorm_robust}
\end{subtable}

\vspace{0.5em}

\begin{subtable}{\textwidth}
\centering
\begin{tabular}{l r|ccccccc}
\toprule
RevRIN & \textbf{Diagnosis} &  \texttt{Chebyshev} &   \texttt{Manhattan} & \texttt{Frobenius} &   \texttt{Spectral}  &   \texttt{Nuclear}  &  \texttt{Wass-0} &   \texttt{Wass-1}  \\
\midrule
\multirow{2}{*}{No} 
    & MCI+IMP & 0.7335\,$\pm$\,0.1450 & 0.6837\,$\pm$\,0.1102 & 0.7221\,$\pm$\,0.2158 & 0.7393\,$\pm$\,0.1506 & 0.6176\,$\pm$\,0.0846 & 0.6708\,$\pm$\,0.0860 & 0.7060\,$\pm$\,0.1183 \\
    & Normal  & 0.5708\,$\pm$\,0.1293 & 0.7543\,$\pm$\,0.2452 & 0.6270\,$\pm$\,0.2297 & 0.6041\,$\pm$\,0.3192 & 0.7686\,$\pm$\,0.2488 & 0.6551\,$\pm$\,0.0409 & 0.5908\,$\pm$\,0.0899 \\
\multirow{2}{*}{Yes} 
    & MCI+IMP & 0.7164\,$\pm$\,0.1543 & 0.6864\,$\pm$\,0.1087 & 0.7290\,$\pm$\,0.1660 & 0.7150\,$\pm$\,0.1406 & 0.7415\,$\pm$\,0.1576 & 0.6711\,$\pm$\,0.0813 & 0.6814\,$\pm$\,0.1155 \\
    & Normal  & 0.6022\,$\pm$\,0.1475 & 0.6382\,$\pm$\,0.1027 & 0.5800\,$\pm$\,0.1619 & 0.6099\,$\pm$\,0.1526 & 0.5635\,$\pm$\,0.1480 & 0.6447\,$\pm$\,0.1253 & 0.6409\,$\pm$\,0.1488 \\
\bottomrule
\end{tabular}
\caption{GPT-2}
\label{tab:mae_gpt2_instancenorm_robust}
\end{subtable}

\vspace{0.5em}

\begin{subtable}{\textwidth}
\centering
\begin{tabular}{l r|ccccccc}
\toprule
RevRIN & \textbf{Diagnosis} &  \texttt{Chebyshev} &   \texttt{Manhattan} & \texttt{Frobenius} &   \texttt{Spectral}  &   \texttt{Nuclear}  &  \texttt{Wass-0} &   \texttt{Wass-1}  \\
\midrule
\multirow{2}{*}{No} 
    & MCI+IMP & 0.8999\,$\pm$\,0.2174 & 0.8091\,$\pm$\,0.2217 & 0.9315\,$\pm$\,0.2640 & 0.8965\,$\pm$\,0.2076 & 0.9229\,$\pm$\,0.2648 & 0.8902\,$\pm$\,0.2227 & 0.8962\,$\pm$\,0.2140 \\
    & Normal  & 0.3627\,$\pm$\,0.1691 & 0.3647\,$\pm$\,0.1737 & 0.3636\,$\pm$\,0.1699 & 0.3729\,$\pm$\,0.1629 & 0.3559\,$\pm$\,0.1735 & 0.3693\,$\pm$\,0.1806 & 0.3706\,$\pm$\,0.1380 \\
\multirow{2}{*}{Yes} 
    & MCI+IMP & 0.8949\,$\pm$\,0.2164 & 0.8971\,$\pm$\,0.2222 & 0.9034\,$\pm$\,0.2344 & 0.8983\,$\pm$\,0.2210 & 0.8902\,$\pm$\,0.2126 & 0.8955\,$\pm$\,0.2250 & 0.8876\,$\pm$\,0.1889 \\
    & Normal  & 0.3673\,$\pm$\,0.1671 & 0.3696\,$\pm$\,0.1806 & 0.3619\,$\pm$\,0.1864 & 0.3660\,$\pm$\,0.1742 & 0.3752\,$\pm$\,0.1724 & 0.3693\,$\pm$\,0.1806 & 0.3706\,$\pm$\,0.1380 \\
\bottomrule
\end{tabular}
\caption{BERT}
\label{tab:mae_bert_instancenorm_robust}
\end{subtable}

\caption{MAE ($\pm$ std) with/without RevRIN}
\label{tab:mae_all_instancenorm_robust_ablation}
\end{table*}

\end{document}